\documentclass{article} % For LaTeX2e
\usepackage{iclr2021_conference,times}

\usepackage{amsmath,amsfonts,bm}

\def\eqref#1{equation~\ref{#1}}
\def\1{\bm{1}}

\DeclareMathAlphabet{\mathsfit}{\encodingdefault}{\sfdefault}{m}{sl}
\SetMathAlphabet{\mathsfit}{bold}{\encodingdefault}{\sfdefault}{bx}{n}

\usepackage{graphicx}
\usepackage{hyperref}
\hypersetup{
    colorlinks,
    linkcolor={purple!50!black},
    citecolor={green!50!black},
    urlcolor={purple!30!black}
}
\usepackage{url}
\usepackage[font={small}]{caption}

\title{Towards continual task learning in artificial neural networks: current approaches and insights from neuroscience}

\author{{David McCaffary}
\\
Wellcome Centre for Integrative Neuroimaging\\
University of Oxford\\
\texttt{david.mccaffary@magd.ox.ac.uk} \\
}

\iclrfinalcopy % Uncomment for camera-ready version, but NOT for submission.
\begin{document}

\maketitle

\begin{abstract}
The innate capacity of humans and other animals to learn a diverse, and often interfering, range of knowledge and skills throughout their lifespan is a hallmark of natural intelligence, with obvious evolutionary motivations. In parallel, the ability of artificial neural networks (ANNs) to learn across a range of tasks and domains, combining and re-using learned representations where required, is a clear goal of artificial intelligence. This capacity, widely described as continual learning, has become a prolific subfield of research in machine learning. Despite the numerous successes of deep learning in recent years, across domains ranging from image recognition to machine translation, such \textit{continual} task learning has proved challenging. Neural networks trained on multiple tasks in sequence with stochastic gradient descent often suffer from representational interference, whereby the learned weights for a given task effectively overwrite those of previous tasks in a process termed \textit{catastrophic forgetting}. This represents a major impediment to the development of more generalised artificial learning systems, capable of accumulating knowledge over time and task space, in a manner analogous to humans. A repository of selected papers and implementations accompanying this review can be found at \url{https://github.com/mccaffary/continual-learning}.
\end{abstract}

\section*{Introduction and historical context}

Several definitions of intelligence have emerged in machine learning literature in recent years \citep{legg2007universal, tenenbaum2011grow}. Although many of these differ in particularities, one consistent requirement is that an intelligent agent must have the capacity to accumulate and maintain task proficiency across experiences \citep{hinton1986learning, hassabis2017neuroscience}. In contrast to biological agents, existing neural network approaches have proved lacking in this regard, with sequential task learning often resulting in catastrophic forgetting of previously encountered tasks \citep{french1999catastrophic,mccloskey1989catastrophic,ratcliff1990connectionist}. A proliferation of research seeking to alleviate the catastrophic forgetting problem has emerged in recent years, motivated by the requirement for machine learning pipelines to accumulate and analyse vast data streams in real-time \citep{parisi2019continual, hadsell2020embracing}. Despite significant progress being made through such research, both in theory and application, the sub-field of continual learning research is vast, and therefore benefits from clarification and unifying critical appraisal. Simple categorisation of these approaches according to network architecture, regularisation, and training paradigm proves useful in structuring the literature of this increasingly important sub-field. Furthermore, many existing approaches to alleviating catastrophic forgetting in neural networks draw inspiration from neuroscience \citep{hassabis2017neuroscience}. In this review, both of these issues will be addressed, providing a broad critical appraisal of current approaches to continual learning, while interrogating the extent to which insight might be provided by the rich literature of learning and memory in neuroscience.

\section*{Architectural considerations}
\label{gen_inst}

The influence of network architecture on task performance has been widely described in machine learning \citep{lecun2015deep}, and represents a fruitful area of continual learning research \citep{parisi2019continual,kemker2017fearnet}. In particular, attention has focused on network architectures which dynamically reconfigure in response to increasing training data availability, primarily by recruiting the training of additional neural network units or layers.

\textit{Progressive neural networks} are one such example, where a dynamically expanding neural network architecture is employed \citep{rusu2016progressive}. For training on each subsequent task, this model recruits additional neural networks which are trained on these tasks, while transfer of learned knowledge across tasks is facilitated by learned ‘lateral’ connections between the constituent networks (Figure 1A). Together, this alleviates catastrophic forgetting across a range of reinforcement learning benchmarks, such as Atari games, and compares favourably with baseline methods which leverage pre-training or fine-tuning of model parameters (Figure 1B,C). Despite the empirical successes of progressive networks, an obvious conceptual limitation is that the number of network parameters grows with the number of tasks experienced. For sequential task training (on \textit{n} tasks), the broader applicability of this method, as $n\to\infty$, remains unclear.

\begin{figure}[h]
\includegraphics[width=14cm]{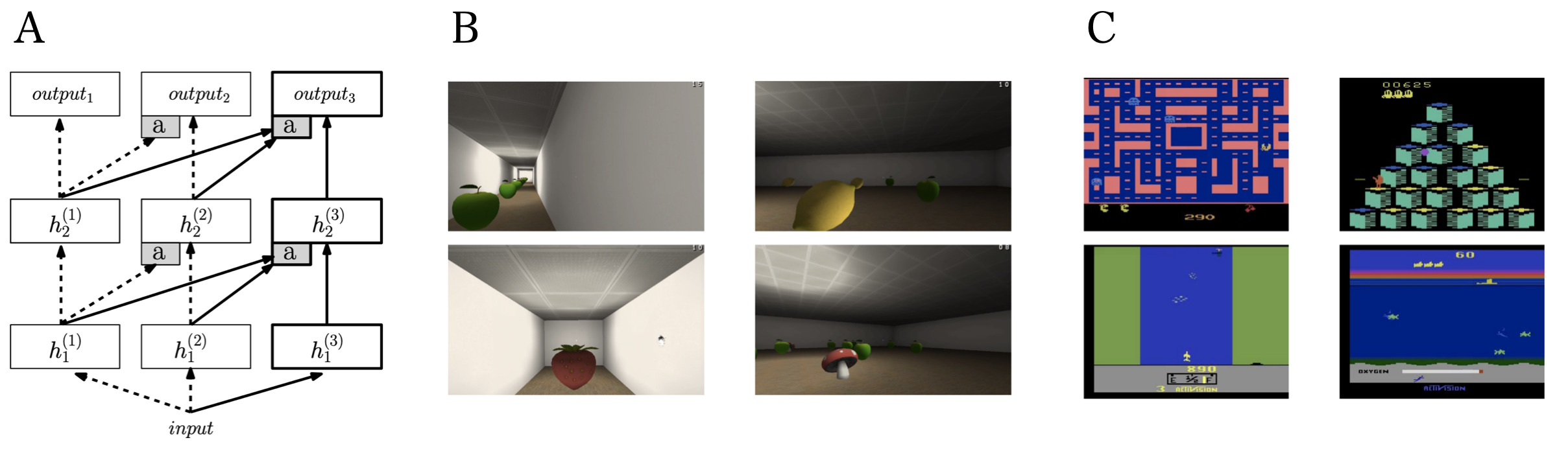}
\caption{\textbf{A)} Schematic of the Progressive neural network architecture, where each of the three columns represents a constituent neural network instantiated to solve a given task. Concretely, the first two columns represent networks trained on tasks 1 and 2, respectively. The third column on the right represents a third network added to solve a novel task, and can draw on previously learned features via lateral connections between the networks (as indicated by the arrows; adapted from \citep{rusu2016progressive}). \textbf{B)} An example task domain (adapted from \citet{rusu2016progressive}), in which a range of 3-dimensional labyrinths are navigated to attain reward. The structural diversity of these environments makes them an ideal test of continual task learning. \textbf{C)} Atari 2600 games similarly offer a paradigm for testing continual learning, but with a richer and more complex set of tasks and environments \citep{rusu2016progressive}. }
\end{figure}

An alternative approach was proposed by \citet{draelos2017neurogenesis}, which takes direct inspiration from hippocampal neurogenesis, a well-described phenomenon from neuroscience in which new neurons are formed in adulthood, raising intriguing questions of learning and plasticity in both natural and artificial intelligence settings \citep{aimone2014regulation}. The method proposed in this paper, termed \textit{neurogenesis deep learning}, is conceptually similar to that of progressive neural networks, but in this instance involves additional neurons in deep neural network layers being recruited as the network is trained on subsequent tasks. \citet{draelos2017neurogenesis} implement this as an autoencoder trained on the MNIST dataset of handwritten digits. As a greater range of digits is added incrementally to the training distribution, units are added in parallel to the autoencoder, thereby giving rise to the ‘neurogenesis’ in this dynamic network architecture. The autoencoder network in this instance preserves weights associated with previously learned tasks using a form of replay, while the reconstruction error provides an indication of how well representations of previously learned digits are preserved across learning of subsequent digits (that is, subsequent task learning). This paper presents an elegant idea for mitigating catastrophic forgetting, but further experiments are required to fully appraise its potential. For instance, the incremental training data used in this paper is solely in the form of discrete, one-hot categories, rather than the more challenging (and more naturalistic) scenario of novel data accumulating gradually, or without clear boundaries.

Both of the approaches discussed so far have involved dynamic network architectures, but nonetheless ones in which networks or units are recruited and incorporated in response to subsequent tasks. An alternative method has been advanced by \citet{cortes2017adanet}, in which no network architecture is explicitly encoded. Instead, the proposed AdaNet algorithm adaptively selects both the optimal network architecture and weights for the given task. When tested on binary classification tasks drawn from the popular CIFAR-10 image recognition dataset, this approach performed well, with the algorithm automatically learning appropriate network architectures for the given task. Although AdaNet has not been tested exhaustively in the context of continual learning, it represents an appealing method of dynamically reconfiguring the network to mitigate catastrophic forgetting with subsequent tasks. Overall, some combination of these approaches – a dynamic network architecture and an algorithm for automatically inferring the optimal architecture for newly encountered tasks – might offer novel solutions to the continual learning problem.

\section*{Regularisation}
\label{headings}

Imposing constraints on the neural network weight updates is another major area of continual learning research \citep{goodfellow2013empirical}. Such regularisation approaches have proved popular in recent years, and many derive inspiration from models of memory consolidation in theoretical neuroscience \citep{fusi2005cascade, losonczy2008compartmentalized}.

\subsection*{Learning without forgetting}

\textit{Learning without forgetting} (LwF) is one such proposed regularisation method for continual learning \citep{li2017learning} and draws on knowledge distillation \citep{hinton2015distilling}. Proposed by Hinton and colleagues, knowledge distillation is a technique in which the learned knowledge from a large, regularised model (or ensemble of models) is distilled into a model with many fewer parameters (the details of this technique, however, are beyond the scope of this work). This concept was subsequently employed in the LwF algorithm to provide a form of functional regularisation, whereby the weights of the network trained on previous tasks or training data are enforced to remain similar to the weights of the new network trained on novel tasks. Informally, LwF aims to effectively take a representation of the network before training on new tasks. In \citet{li2017learning}, this was implemented as a convolutional neural network, in which only novel task data was used to train the network, while the ‘snapshot’ of the prior network weights preserved good performance on previous tasks. This approach has garnered significant attention in recent years, and offers a novel perspective on the use of knowledge distillation techniques in alleviating catastrophic forgetting. However, Learning without Forgetting has some notable limitations. Firstly, it is highly influenced by task history, and is thus susceptible to forming sub-optimal representations for novel tasks. Indeed, balancing stability of existing representations with the plasticity required to efficiently learn new ones is a major unresolved topic of research in continual learning. A further limitation of LwF is that, due to the nature of the distillation protocol, training time for each subsequent task increases linearly with the number of tasks previously learned. For broad applicability, this practically limits the capacity of this technique to handle pipelines of training data for which novel tasks are encountered regularly.

\subsection*{Elastic weight consolidation}

In recent years, one of the most prominent regularisation approaches to prevent catastrophic forgetting is that of \textit{elastic weight consolidation} (EWC) \citep{kirkpatrick2017overcoming}. EWC, which is suitable for supervised and reinforcement learning paradigms, takes direct inspiration from neuroscience, where synaptic consolidation is thought to preserve sequential task performance by consolidating the most important features of previously encountered tasks \citep{yang2009stably}. Intuitively, EWC works by slowing learning of the network weights which are most relevant for solving previously encountered tasks. This is achieved by applying a quadratic penalty to the difference between the parameters of the prior and current network weights, with the objective of preserving or \textit{consolidating} the most task-relevant weights. It is this quadratic penalty, with its ‘elastic’ preservation of existing network weights, which takes inspiration from synaptic consolidation in neuroscience, and is schematically represented in Figure 2A. More formally, the loss function of EWC, $L(\theta)$, is given by:
$$
\mathcal{L}(\theta)=\mathcal{L}_{B}(\theta)+\sum_{i} \frac{\lambda}{2} F_{i}\left(\theta_{i}-\theta_{A, i}^{*}\right)^{2}
$$
Where $\theta$ represents the parameters of the network, $L_B(\theta)$ represents the loss for task \textbf{B}, $\lambda$ is a hyperparameter indicating the relative importance of previously encountered tasks compared to new tasks, $F$ is the Fisher information matrix, and finally $\theta A^*$ represents the trainable parameters of the network important for solving previously encountered tasks. Intuitively, this loss function can be understood as penalising large differences between previous and current network weights (the term within the brackets). In EWC, the Fisher information matrix is used to give an estimation of the importance of weights for solving tasks, by using an importance weighting proportional to the diagonal of the Fisher information metric over the old parameters for the previous task. While conceptually elegant, this presents a notable limitation of EWC: exact computation of the Fisher diagonal has complexity linear with the number of outputs, limiting the applicability of this method to low-dimensional output spaces.

Empirically, EWC performs well on supervised and reinforcement learning tasks, such as MNIST digit classification and sequential Atari 2600 games, respectively. However, the suitability of the quadratic penalty term (which is only derived for the two-task case in the original paper) has been questioned for cases of multi-task learning \citep{huszar2018note}. Additionally, building on the strong empirical results presented in the paper, \citet{kemker2018measuring} demonstrated in further experiments that EWC fails to learn new classes incrementally in response to accumulating training data (Figure 2C). Overall, EWC is a promising method for alleviating catastrophic forgetting, but several details regarding its broader applicability (and theoretical underpinning) remain unresolved.

\begin{figure}[h]
\includegraphics[width=14cm]{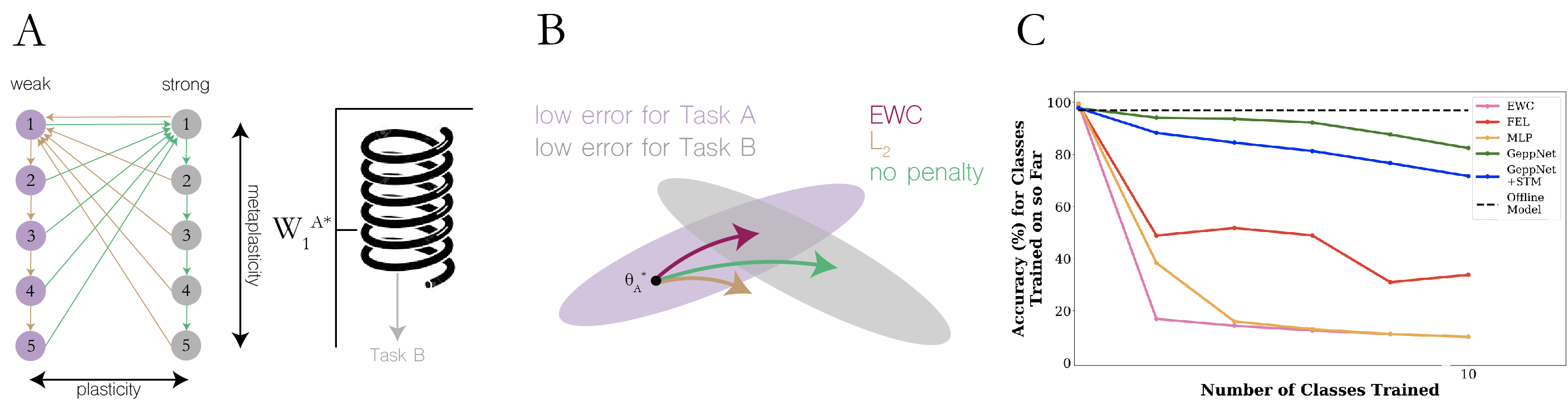}
\caption{\textbf{A)} Schematic of the analogy between synaptic consolidation (left) and the regularisation of EWC (right), in which network weights important for solving previous tasks are preserved in an ‘elastic’ manner. Adapted from \citet{hassabis2017neuroscience}. \textbf{B)} Schematic showing the parameter spaces of tasks A and B, for which EWC finds the optimal balance of weights for preserving performance of task A after training on task B. Unlike other regularisation approaches (such as L2 regularisation, as depicted), EWC does so by explicitly calculating the importance of weights in the network for solving a given task. Adapted from \citet{kirkpatrick2017overcoming}. \textbf{C)} Atari 2600 games similarly offer a paradigm for testing continual learning, but with a richer and more complex set of tasks and environments \citep{kemker2018measuring}.}
\end{figure}

\subsection*{Synaptic intelligence}

An approach closely related to EWC is that of \textit{synaptic intelligence} \citep{zenke2017continual}. In this method, however, individual synapses (the connections between neurons or units) estimate their importance in solving a given task. Such \textit{intelligent synapses} can then be preserved for subsequent tasks by penalising weight updates, thereby mitigating catastrophic forgetting of previously learned tasks. Intuitively, synaptic intelligence can be considered a mechanism of anchoring network weights relevant for previous tasks to their existing values, and decelerating updates of these weights to prevent over-writing of previous task performance.

In summary, numerous regularisation methods have been developed to aid continual learning. By modulating gradient-based weight updates, these methods aim to preserve the performance of the model across multiple tasks trained in sequence. Many such regularisation methods have garnered interest from the research community, both due to both their theoretical appeal and strong empirical validation. Ultimately, however, none of these approaches has offered a comprehensive solution to the continual learning problem, and it is likely that a deeper understanding of credit assignment within neural networks will drive this research further.

\section*{Training regimes}

Beyond the model itself, the training regime employed is critical to sequential task performance, and represents a rich avenue of continual learning research. Although numerous training paradigms have been described, this review will focus on those most directly developed to alleviate catastrophic forgetting.

\subsection*{Transfer learning}

Historically, limitations of dataset size have motivated the study of \textit{transfer learning} for machine learning systems, with the aim of initially training ANNs on large datasets before transferring the trained network parameters to other tasks \citep{bengio2012practical,higgins2016early}. The recent successes of transfer learning in fields such as computer vision and natural language processing are well-documented. For instance, \citet{yosinski2014transferable} demonstrated the efficacy of transfer learning for computer vision in a range of benchmarking datasets. It was then found that subsequent layers in the hierarchy of the neural network display representational features which are reminiscent of human visual cognition (such as edge detection in lower layers and image-specific feature detection in higher layers).

Successes of transfer learning in few-shot learning paradigms, which aim to perform well upon first presentation of a novel task, have been well-described in recent years \citep{palatucci2009zero,vinyals2016matching}. However, translation of this potential into alleviating catastrophic forgetting has proved more challenging. One of the earliest attempts to leverage transfer for continual learning was implemented in the form of a hierarchical neural network, termed CHILD, trained to solve increasingly challenging reinforcement learning problems \citep{ring1998child}. This model was not only capable of learning to solve complex tasks, but also demonstrated a degree of continual learning – after learning nine task structures, the agent could still successfully perform the first task when returned to it. The impressive (and perhaps overlooked) performance of CHILD draws on two main principles: firstly, transfer of previously learned knowledge to novel tasks; secondly, incremental addition of network units as more tasks are learned. In many ways, this model serves as a precursor of progressive neural networks \citep{rusu2016progressive}, and offers an appealing account of transfer learning in aiding continual task learning.

\subsection*{Curriculum learning}

In parallel to this, \textit{curriculum learning} has gained attention of improving continual learning capacities in ANNs \citep{bengio2009curriculum,graves2016hybrid}. Broadly, curriculum learning can be defined as the phenomenon by which both natural and artificial intelligence agents learn more efficiently when the training dataset contains some inherent structure, and when exemplars provided to the learning system are organised in a meaningful way (for instance, progressing in difficulty) \citep{bengio2009curriculum}. When trained on datasets such as MNIST, curriculum learning both accelerates the learning process (as measured by time or training steps to reach the global minimum) and helps prevent catastrophic forgetting \citep{bengio2009curriculum}. However, one limitation of curriculum learning in ANNs is the assumption that task difficulty can be represented in a linear and uniform manner (often described as a ‘single axis of difficulty’), disregarding the nuances of each task structure. Nevertheless, curriculum learning is a promising, and underexplored, avenue of research for better continual learning performance in neural networks.

\subsection*{Generative replay}

The neural phenomenon of hippocampal (or, more generally, memory) replay has recently garnered attention with respect to the design of ANNs \citep{skaggs1996replay,shin2017continual,kemker2018measuring}. \citet{shin2017continual} designed the model most obviously inspired by hippocampal replay, with a training regime termed \textbf{Deep Generative Replay}. This comprises a generative model and a task-directed model, the former of which is used to generate representative data from previous tasks, from which a sample is selected and interspersed with the dataset of the new task. In this way, the model mimics hippocampal replay, drawing on statistical regularities from previous experiences when completing novel tasks \citep{liu2019human}. This approach, which shares some conceptual similarities to the model proposed by \citet{draelos2017neurogenesis}, displays a substantial improvement in continual learning compared to complexity-matched models lacking replay.

Several other implementations of replay in neural networks have also been described, including a straightforward experience replay buffer of all prior events for a reinforcement learning agent \citep{rolnick2018experience}. This method, called \textbf{CLEAR}, attempts to address the stability-plasticity tradeoff of sequential task learning, using off-policy learning and replay-based behavioural cloning to enhance stability, while maintaining plasticity via on-policy learning. This outperforms existing deep reinforcement learning approaches with respect to catastrophic forgetting, but might prove unsuitable in cases where storage of a complete memory buffer is intractable.

\begin{figure}[ht]
\includegraphics[width=14cm]{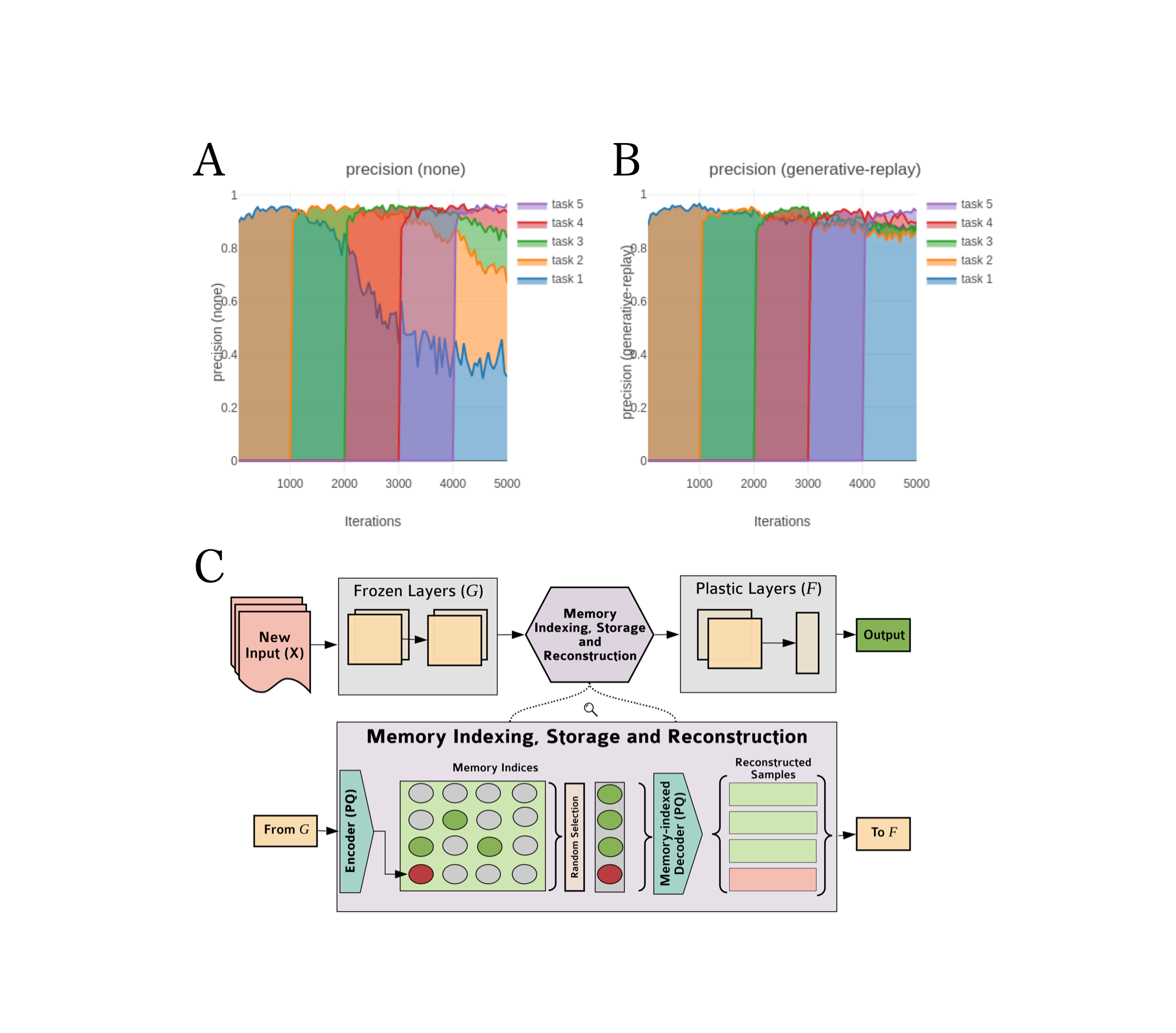}
\caption{Sub-panels \textbf{A} and \textbf{B} demonstrate the efficacy of deep generative replay in alleviating catastrophic forgetting. \textbf{A)} When a neural network is trained on sequential tasks (in this instance, from the permuted MNIST dataset) with vanilla gradient descent, catastrophic forgetting of previous task performance occurs due to overwriting of the weights associated with these prior tasks (\url{https://github.com/kuc2477/pytorch-deep-generative-replay}). As multiple tasks in sequence are encountered, performance on previous tasks can decrease dramatically, as the network weights optimised for these prior tasks are ‘catastrophically forgotten’. \textbf{B)} Conversely, when the network is trained with deep generative replay, continual learning across tasks is much improved. By sampling training data from previous tasks and interleaving this with the current task, this method enables multiple tasks to be learned in sequence with the same network, without catastrophic forgetting of earlier task performance (adapted from the PyTorch implementation of deep generative replay found at \url{https://github.com/kuc2477/pytorch-deep-generative-replay}). \textbf{C)} Schematic of the REMIND model \citep{hayes2020remind}, which proposes the replay of compressed data representations in a more biologically plausible framework. REMIND takes an input image (denoted $X$ in the schematic), and compresses this into low-dimensional tensor representations of that training data by passing it through the neural network layers labelled $G$. These compressed representations can then be efficiently stored in memory (as shown by the lower box in the schematic), and indexed for replay as required. The compressed tensor representations can be reconstructed by the neural network layers labelled $F$, and interleaved with current training data.}
\end{figure}

\subsection*{REMIND}

Drawing on the successes of replay in alleviating catastrophic forgetting, \citet{hayes2020remind} proposed a novel method which more accurately reflects the role of memory replay as described by modern neuroscience, implemented as a convolutional neural network. Standard replay approaches for convolutional neural networks (CNNs) rely on presenting raw pixel input from previously encountered data interleaved with novel data. Although effective, this approach is both biologically implausible and memory-intensive. By contrast, the REMIND (\textbf{re}play using \textbf{m}emory \textbf{ind}exing) algorithm proposed in this paper replays a \textit{compressed} representation of previously encountered training data, thereby enabling efficient replay and online training. This approach is directly inspired by hippocampal indexing theory \citep{teyler2007hippocampal}, and proposes a solution to the issue with prior replay implementations of having to store raw training data from previous tasks. In REMIND, this compression of input data is achieved using layers of the neural network, where the raw input data is compressed into a lower-dimensional tensor representation (for instance, in the context of CNNs, a feature map) for replay (see Figure 3C). Hayes et al. implement this using product quantization, a technique for compressing data which has a significantly lower reconstruction error compared to other methods (the details of product quantization are beyond the scope of this review, but see \citet{jegou2010product} for a comprehensive account). This compression proves highly effective in maximising memory efficiency: REMIND can store one million compressed representations compared to just 20,000 when raw data input is stored in alternative models, matched for memory capacity. Empirically, replay of compressed training data was shown to confer strong benefits, whereby REMIND outperforms constraint-matched methods on incremental class learning tasks derived from the ImageNet dataset \citep{deng2009imagenet}.

\section*{Can inspiration be drawn from neuroscience?}

Neuroscience (in particular, cognitive neuroscience) and artificial intelligence have long been intertwined both in aetiology and research methodologies \citep{hassabis2017neuroscience,hinton1986learning}, and this intersection has already inspired numerous approaches to continual learning research. Many of the methods described previously in this review draw inspiration from neuroscience, either implicitly or explicitly (for instance, generative replay, the REMIND algorithm, and transfer learning are all conceptually indebted to decades of neuroscience research).

One line of justification for this approach is that there are studies demonstrating a phenomenon analogous to catastrophic forgetting in humans, suggesting that a shared mechanistic framework might underlie continual learning (and its inherent limitations) in both humans and ANNs \citep{pallier2003brain,mareschal2007neuroconstructivism}. The first of these studies, \citet{pallier2003brain}, examined language acquisition and overwriting in Korean-born subjects whose functional mother tongue was French (due to being adopted before the age of 8), and had no conscious knowledge of the Korean language, as verified by behavioural testing. Functional neuroimaging (fMRI) demonstrated that the Korean-born francophone subjects displayed no greater (cortical) response to the Korean language in the setting of passive listening compared to French subjects with no exposure to Korean. This was interpreted as a form of over-writing, or catastrophic forgetting, of the first language by the second. The significance of these results is unclear, particularly given the limited literature on human catastrophic forgetting, but represents an interesting mandate for the use of cognitive neuroscience as a source of inspiration for continual learning research.

\subsection*{Replay in brains and neural networks}

From the perspective of neuroscience, several mechanistic features underpinning human continual learning have been dissected, such as memory replay (perhaps as a means of transferring learned knowledge from short-term to long-term storage), curriculum and transfer learning paradigms, structural plasticity, and the integration of multiple sensory modalities to provide rich sensory context for memories \citep{parisi2019continual}.

One phenomenon widely considered to contribute to continual learning is memory replay or hippocampal replay, defined as the re-activation (hence, replay) of patterns of activity in hippocampal neurons during states of slow-wave sleep and passive, resting awake \citep{skaggs1996replay,dave2000song,rudoy2009strengthening}. Such replay episodes are thought to provide additional trials serving to rehearse task learning and generalise knowledge during so-called ‘offline’ learning, and were first identified by recording hallmarks of brain activity during learning and mapping these onto covarying activity patterns identified during sleep \citep{rasch2018memory}. An elegant demonstration of this phenomenon in humans was provided by \citet{rudoy2009strengthening}, whereby subjects learned the position of arbitrary objects on a computer screen, with each object presented in association with a unique and characteristic sound. The participants then slept for a short period of time, with electroencephalography (EEG) used to identify different stages of sleep. During slow-wave sleep, the characteristic sounds for half of the objects were played at an audible but unobtrusive volume. It was found that the participants subsequently recalled the positions of these sound-consolidated objects with greater accuracy. Replay approaches in machine learning have already proved fruitful, and studies such as these from neuroscience only serve to further motivate replay as a topic of continual learning research.

Replay, as well as the lesser-understood hippocampal pre-play (whereby hippocampal neurons display what is thought to represent simulated activity which can be mapped onto future environments) might aid continual learning through this ‘offline’ consolidation \citep{dragoi2011preplay,bendor2016does}. Although the mechanism remains incompletely described, it has been proposed that replay could contribute to continual learning by promoting the consolidation of previous task knowledge \citep{olafsdottir2018role}. In some ways, this can be considered analogous to pre-training neural networks with task-relevant data, a technique already demonstrated to confer considerable advantages \citep{erhan2010does} . Indeed, the value of neuroscience-inspired research was recently underlined by a study from \citet{van2020brain}. Here, a more biologically plausible form of replay was implemented, whereby instead of storing previous task data, a learned generative model was trained to replay internal, compressed representations of that data. These internal representations for replay were then interleaved with current training data in a context-dependent manner, modulated by feedback connections. This approach also overcomes a potential issue with existing replay methods in machine learning – cases where storing previously encountered data is not permissable due to safety or privacy concerns. Replay of compressed representations, or replay in conjunction with some form of federated learning \citep{kairouz2019advances}, might offer a solution in these instances.

%include something about replay for credit assignment as a new replay hypothesis from neuroscience - cite paper

\subsection*{Complementary learning systems for continual learning}

Complementary learning systems (CLS) theory, first advanced by \citet{mcclelland1995there} and recently updated by \citet{kumaran2016learning}, delineates two distinct structural and functional circuits underlying human learning \citep{mcclelland1995there,kumaran2016learning,girardeau2009selective}. The hippocampus serves as the substrate for short-term memory and the rapid, ‘online’ learning of knowledge relevant to the present task; in parallel, the neocortex mediates long-term, generalised memories, structured over experience. Transfer of knowledge from the former to the latter occurs with replay, and it is intuitive that the catastrophic forgetting of previously learned knowledge in machine learning systems could be mitigated to some extent by such complementary learning systems. This has proved an influential theory in neuroscience, offering an account of the mechanisms by which humans accumulate task performance over time, and has started to provide ideas for how complementary learning systems might aid continual learning in artificial agents.

\subsection*{Transfer and curriculum learning}

Furthermore, the learning strategies employed by humans, of which transfer learning and curriculum learning have come into recent focus, are themselves likely to contribute further to continual learning \citep{barnett2002and,holyoak1997analogical}. Humans are capable of transferring knowledge between domains with minimal interference, and this capability derives from both continual task learning and generalisation of previously learned knowledge. A human learning to play tennis, for instance, can generalise some features of this task learning to a different racquet sport \citep{goode1986contextual}. Such transfer learning is poorly understood at the level of neural computations, and has perhaps been neglected by the neuroscience research community until recently, when attempts to endow artificial agents with these abilities has re-focussed attention on the mechanisms underpinning transfer learning in humans \citep{weiss2016survey}.

Attempts to explain this abstract transfer of generalised knowledge in humans have themselves recapitulated many features of continual learning \citep{doumas2008theory,barnett2002and,pan2009survey,holyoak1997analogical}. For instance, \citet{doumas2008theory} proposed that this is achieved by the neural encoding of relational information between objects comprising a sensory environment. Critically, such relational information would be invariant to nuances and specific features in these objects, and this could aid continual learning by providing a generalised task learning framework. Although the neural coding for such a relational framework has not yet been elicited, an intriguing recent paper by \citet{constantinescu2016organizing} proposed abstract concepts are encoded by grid cells in the human entorhinal cortex in a similar way to maps of physical space. Just as replay and memory consolidation have already led approaches to continual learning, such research might inspire novel approaches to alleviating catastrophic forgetting.

A related, and perhaps lesser-studied, learning paradigm is that of curriculum learning \citep{bengio2009curriculum,elman1993learning}. Intuitively, curriculum learning states that agents (both natural and artificial) learn more effectively when learning examples are structured and presented in a meaningful manner. The most obvious instantiation of this is to increase the difficulty of the learning rules throughout the sequence of examples presented; indeed, this is consistent with the structure of most human educational programmes \citep{krueger2009flexible,goldman1995complexity}. It has been appreciated for some time that this form of non-random learning programme aids human continual learning \citep{elman1993learning,krueger2009flexible}; however, the theoretical underpinnings of this are only starting to be elicited. Curriculum learning has the potential to enhance continual learning in neural networks by providing more structured training regimes, which emphasise the features of the training dataset which are most relevant to the tasks. Ultimately, however, more work is required to explore the promise of this.

\subsection*{Multi-sensory integration and attention}

It has been appreciated for some time in the field of cognitive neuroscience that humans receive a stream of rich multi-sensory input from the environment, and that the ability to integrate this information into a multi-modal representation is critical for cognitive functions ranging from reasoning to memory \citep{spence2010crossmodal,spence2014orienting,stein1993merging,stein2014development}. By enriching the context and representation of individual memories, multi-sensory integration is thought to aid human continual learning.

There is also evidence that attention is a cognitive process contributing to continual learning in humans \citep{flesch2018comparing}. Here, when the task learning curriculum was designed in a manner permitting greater attention (namely, with tasks organised in blocks for training, rather than ‘interleaved’ task training), continual learning of the task in the (human) subjects was enhanced. Even if optimal training regimes differ across biological and artificial agents, this underlines the importance of curriculum and attention in addressing catastrophic forgetting.

\section*{Future perspective: bridging neuroscience and machine learning to inspire continual learning research}

The inherent efficacy of human continual learning and its cognitive substrates is perhaps most impressive when contrasted with the current inability to endow AI agents with similar properties. With a projected increase in global data generation from 16 zettabytes annually in 2018 to over 160 zettabytes annually by 2025 (and the consequent intractability of comprehensive storage), there is a clear motivation for developing machine learning systems capable of continual learning in a manner analogous to humans (IDC White Paper, 2017; Tenenbaum et al., 2011).

The super-human performance of deep reinforcement learning agents on a range of complex learning tasks, from Atari 2600 video games to chess, has been well-publicised in recent years \citep{mnih2015human,silver2016mastering,kasparov2018chess,lecun2015deep}. However, these successes conceal a profound limitation of such machine learning systems: an inability to sustain performance when trained sequentially over a range of tasks. Traditionally, approaches to the issue of catastrophic forgetting have focussed on training regime, and often remain tangential to the cause of such forgetting. In the future, bridging the conceptual gap between continual learning research and the rich literature of learning and memory in neuroscience might prove fruitful, as motivated by several of the examples already discussed in this review.

\subsection*{Parallels of CLS theory in machine learning}

For example, with respect to CLS theory, biologically inspired neural network architectures involving two different network parameters (a ‘plastic’ parameter for slow-changing information, and a rapidly updating parameter) have existed for decades \citep{hinton1987using}. Indeed, these networks outperformed state-of-the-art ANNs in continual learning-related tasks at their time of development. When considered from the perspective of CLS theory, this suggests that the parallel and complementary functions of the hippocampus and neocortex in human memory contribute to continual learning.

More recent models, such as the Differentiable Neural Computer (DNC), also support the view that having complementary memory systems supports continual learning \citep{graves2016hybrid}. The DNC architecture consists of an artificial neural network and an external memory matrix, to which the network has access to store and manipulate data structures (broadly analogous to random-access memory). As such, a DNC can be interpreted as having ‘short-term’ and ‘long-term’ memory proxies, and the capacity to relay information between them. This model is capable of solving complex RL problems, and answering natural language questions constructed to mimic reasoning, lending further support to the contribution of complementary learning systems in human continual learning. More recently, models directly inspired by neuroscience have attempted to study this principle of fast- and slow-learning systems \citep{whittington2020tolman}, but have not yet been more broadly explored through the lens of continual learning.

\subsection*{Emerging significance of multi-sensory integration and attention in artificial intelligence agents for continual learning}

In the context of continual learning in machine learning, multi-sensory integration (often called ‘multi-modal integration’ in this context) has an obvious benefit of conferring additional information from different modalities when the environment is uncertain or has high entropy. Indeed, multi-modal machine learning has demonstrated efficacy in a range of task learning paradigms, such as lip reading, where the presence of both audio (phoneme) and visual (viseme) information improves performance compared to a uni-sensory training approach \citep{ngiam2011multimodal}. Ultimately, greater investigation of multi-modal machine learning could unravel the value of such integration across domains, and offer approaches to aiding continual learning in settings where the environment is unpredictable or multimodal. The role of attention in human continual learning was underlined by a recent study endowing ANNs with a ‘hard attention mask’, an attentional gating mechanism directly inspired by human cognitive function \citep{serra2018overcoming}. This substantially decreased catastrophic forgetting in this model when trained on image classification tasks, thereby emphasising attention as an important contribution to continual learning.

\section*{Conclusion}

Advances in deep learning have accelerated in recent years \citep{krizhevsky2012imagenet, vaswani2017attention}, capturing the imagination of researchers and the public alike with their capacity to achieve superhuman performance on tasks \citep{silver2016mastering}, and aid scientific discovery \citep{senior2020improved}. However, if machine learning pipelines are ever going to dynamically learn new tasks in real time, with interfering goals and multiple input datasets, the continual learning problem must be addressed. In this review, several of the most promising avenues of research have been appraised, with many of these deriving inspiration from neuroscience. Although much progress has been made, no existing approach adequately solves the continual learning problem. This review argues that more directly bridging continual learning research with neuroscience might offer novel insights and inspiration, ultimately guiding the development of novel approaches to catastrophic forgetting which bring the performance of artificial agents closer to that of humans – assimilating knowledge and skills over time and experience.

\section*{Code availability}
A repository of selected papers and implementations accompanying this review can be found at \url{https://github.com/mccaffary/continual-learning}.

%\subsubsection*{Acknowledgments}
%Use unnumbered third level headings for the acknowledgments. All
%acknowledgments, including those to funding agencies, go at the end of the paper.

\bibliography{iclr2021_conference}
\bibliographystyle{iclr2021_conference}

\end{document}